\documentclass[10pt,twocolumn,letterpaper]{article}

\usepackage{times}
\usepackage{epsfig}
\usepackage{graphicx}
\usepackage{amsmath}
\usepackage{amssymb}

\usepackage{subfigure}
\usepackage{pgfplots} 

\usepackage[colorlinks]{hyperref}
\usepackage{multirow}
\usepackage{array}

\usepackage{algorithm}
\usepackage{algorithmic}

\newcommand{\bx}{\boldsymbol{x}}
\newcommand{\bc}{\boldsymbol{c}}

\newcommand{\sY}{\mathcal{Y}}

\newcommand{\sS}{\mathcal{S}}

\DeclareMathOperator*{\argmax}{argmax}

\newcommand{\surdgreen}{\textcolor{green}{$\surd$}}
\newcommand{\approxblu}{\textcolor{blue}{$\approx$}}
\newcommand{\crossred}{\textcolor{red}{$\bigotimes$}}

\newcommand{\field}[1]{\mathbb{#1}}

\newcommand{\R}{\field{R}}

\renewcommand{\Pr}{\field{P}}

\newcommand{\wh}{\widehat}
\newcommand{\ve}{\varepsilon}

\newcommand{\bV}{\boldsymbol{V}}
\newcommand{\yhat}{\wh{y}}

\newlength\figureheightf
\newlength\figurewidthf
\usetikzlibrary{datavisualization}
\tikzset{every picture/.style={font issue={\fontsize{7.5}{10}}},font issue/.style={execute at begin picture={#1\selectfont}}}

\begin{document}
\title{
Active Learning for Online Recognition of Human Activities from Streaming Videos.}

\author{Rocco De Rosa\\
	{\tt\small derosa@dis.uniroma1.it}
	\and
	Ilaria Gori\\
		{\tt\small ilaria.gori@iit.it}
			\and
		Fabio Cuzzolin\\
			{\tt\small fabio.cuzzolin@brookes.ac.uk}
				\and
				Barbara Caputo\\
				{\tt\small caputo@dis.uniroma1.it}
					\and
					Nicol\'{o} Cesa-Bianchi\\
					{\tt\small nicolo.cesa-bianchi@unimi.it}
}

\maketitle

%
\maketitle
\vspace{-0.2cm}
\begin{abstract}
Recognising human activities from streaming videos 
poses unique challenges to learning algorithms: predictive models need to be scalable, incrementally trainable, and must remain bounded in size even when the data stream is arbitrarily long.
Furthermore, as parameter tuning is problematic in a streaming setting, suitable approaches should be parameterless,
and make no assumptions on what class labels may occur in the stream. We present here an approach to the recognition of human actions from streaming data which
meets all these requirements by: (1) incrementally learning a model which adaptively covers the feature space with simple local classifiers; (2) employing an active learning strategy to reduce annotation requests; (3) achieving promising accuracy within a fixed model size. 
Extensive experiments on standard benchmarks show that our approach is competitive with state-of-the-art non-incremental methods, and outperforms the existing active incremental baselines.

\end{abstract}
\vspace{-0.2cm}
\section{Introduction}
\vspace{-0.15cm}
\label{sec:int}
The pervasive presence of cameras in our everyday lives has created a strong demand for automatic methods able to analyse real time video streams. This is especially challenging for videos depicting human activities, as it is the case in TV footages, surveillance cameras, human computer and human robot interactions, and many other applications. The mainstream approaches to action and activity recognition are typically based on an offline training phase (for a review of previous work in activity recognition we refer the reader to Sec.~\ref{sec:rw}). Such a setting leads to several critical issues when dealing with streaming videos:

\smallskip\noindent
\emph{How to continuously learn about activities from the incoming data?} The dynamic nature of the streaming video setting implies that at each time instant new data is made available to the system, which needs to continuously learn from it. This implies refining its model of known human activities and adding new previously unseen activities on the fly.

\smallskip\noindent
\emph{How to minimise the required annotation effort?} Strictly related to the ability to learn on the fly is the issue of how many video fragments should be annotated. While for newly observed activities one might assume that all video fragments should be manually annotated, when analysing footage of known actions only a fraction of the video fragments will likely bring new information, worthy of the annotation effort. In this context, the system should automatically select which fragments are the most informative, and asks for the help of a human annotator in those cases.

\smallskip\noindent
\emph{How to optimise the algorithm's heuristics dynamically?} System's components, such as the features chosen and the algorithm parameters, have a crucial impact on the final performance of any framework. When learning continuously from incoming data, it is difficult to chose these two components optimally, or even just properly. This is because, by the very nature of dynamic data, it is not possible to anticipate what new instances of the known classes the system will face, or what new activities it will be asked to learn.
	

\begin{figure*}[t!]
	\centering
	\includegraphics[width=1\textwidth]{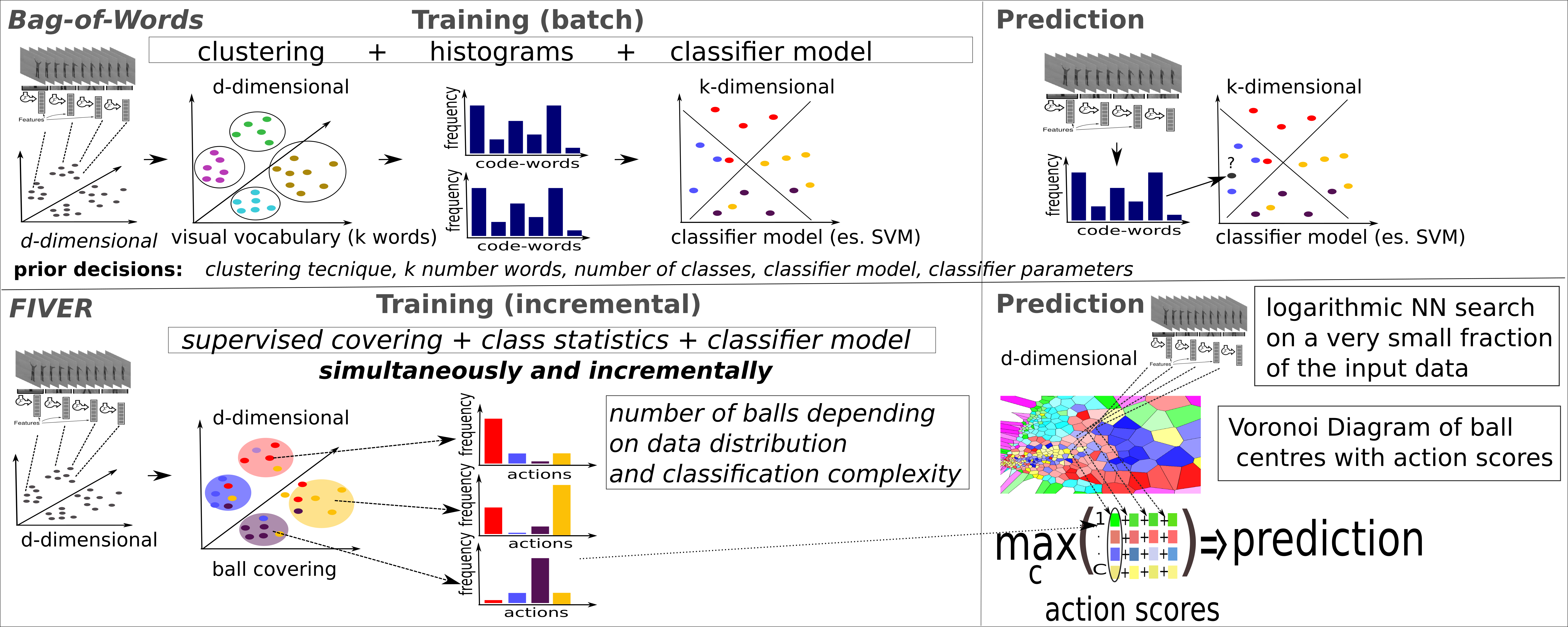}
	\caption{%
		\textbf{Our approach vs Bag-of-Words}. In the classical BoW pipeline~\cite{sivic2003video} (top) vocabulary size, clustering method, and classifier parameters are the result of tuning by cross-validation. In contrast, our approach (see Alg.~\ref{alg:train}) is completely data-driven (nonparametric and parameterless). Our local balls cover the feature space as BoW codewords do, but their number, size, and location dynamically depend on the distribution of feature vectors and associated labels. Rather than having a global classifier act on the entire feature space, a cover of local classifiers associated with updated local class statistics is incrementally built.  An active learning component, based on a suitable confidence measure (see Alg.~\ref{alg:online-budget}), makes it possible to perform active temporal segmentation, leading to continuous activity recognition (Fig.~\ref{fig:spot-gig}).
	}%
	\label{fig:bow-comp} \vspace{-3mm}
\end{figure*}

\smallskip
The contribution of this paper is an algorithm for human activity recognition from streaming videos which, to the best of our knowledge, is the first to address all the challenges listed above in a principled manner. Our starting point is a very recently proposed local algorithm for classification of data streams~\cite{derosa2015abacoc} that is incrementally trainable and nonparametric\footnote{In nonparametric models the model structure is not specified a priori, but determined from data. The implication is not that such models completely lack parameters, but that the number and value of the parameters are flexible and not fixed in advance.}, while providing theoretical guarantees on performance. We leverage on this result, and  extend it to the active learning setting. This leads to a framework that matches all the requirements listed above: (1) it incrementally learns from the incoming data stream, with respect to both known classes and new classes, while being computationally efficient; (2) its active learning component evaluates the informative content of the incoming data with respect to the level of confidence of the system and allows to decide
when the cost of manual annotation is worthwhile; (3) lastly, the nonparametric nature of the approach, combined with its intrinsic locality, allows for fully data-driven learning.

Fig.~\ref{fig:bow-comp} illustrates the workflow of our approach:
(i) each video is associated with an arbitrary number of feature vectors in a feature space; (ii) the incoming training vectors are used to sequentially cover the feature space with balls;
(iii) each ball is associated with an estimate of the conditional class probabilities obtained by collecting statistics around its centre, these estimates are used to make predictions on new unlabeled samples; (iv) the radius of each ball is adjusted accordingly how well each ball predict the new samples around its centre; (v) ball centres are incrementally adjusted to fit the actual data distribution; (vi) the set of balls is organized in a tree structure~\cite{Krauthgamer:2004:NNS:982792.982913}, so that predictions can be computed in time logarithmic in the number of balls.
We call our algorithm Fast active Incremental Visual covERing (FIVER). Extensive experiments on several publicly available databases show that our approach outperforms all existing algorithms for activity recognition from streaming data. Furthermore, we show that by
combining FIVER with the robust temporal segmentation algorithm presented in~\cite{Fanello:2013:KSS:2567709.2567745}
we obtain a system able to deal in a straightforward manner with a realistic continuous active recognition scenario.



\vspace{-0.2cm}
\section{Related Work}
\label{sec:rw}
\vspace{-0.2cm}
Within the vast literature related to action recognition (see~\cite{poppe2010survey} and references therein), research focusing on the streaming setting has gained momentum only recently \cite{gaber2007survey}. Important features required in this context are: \emph{(1) Incremental Updates:} as a large amount of data is generally presented sequentially in a stream, it is desirable for algorithms to update the model adaptively, rather than re-train from scratch.  \emph{(2) Incremental Learning of New Activities:} algorithms should be able to accomodate on the fly new upcoming classes.
\emph{(3) Bounded Size Model:} as the stream could be very large, models should keep a bounded memory footprint, allowing for real-time prediction while avoiding storage issues. This implies the ability to discard useless or old information, which is critical to the tracking of ``drifting concepts''
(i.e., when the optimal decision surface changes over time and the model needs to be rearranged accordingly~\cite{tsymbal2004problem}).
\emph{(4) Data-Driven:} as parameter tuning is problematic in streaming settings, systems with few or no parameters are preferable. \emph{(5) Nonparametric:} as the true structure of the data is only progressively revealed as more examples from the stream are observed, nonparametric algorithms~\cite{opac-b1123996}, which are not committed to any specific family of decision surfaces, are preferable.
\emph{(6) Active Learning:} in a streaming setting, the system needs to learn from each incoming video stream. However, training labels can only be provided by human annotators, who should only be invoked when the system has low confidence in its own prediction of the current label.
\emph{(7) Bounded Request rate:} as querying human annotators is expensive, any practical active learning system for streaming settings should impose a bound on the query rate.

Table~\ref{tab:feat_comp} lists the previous efforts in human activity recognition involving incremental and/or active learning components which, due to their features, are the closest alternatives to our approach.

 \begin{table*}[!tb]
	\begin{center}
\begin{tabular}{|l|c|c|c|c|c|c|c|}
	\hline  & \small{Incr.\ Upd.} &  \small{New Activ.} & \small{Bound.\ Size} & \small{Data Driv.} & \small{Nonpar.} & \small{Act.\ Learn.} & \small{Bound.\ Rate} \\
	\hline \rule{0pt}{4ex} De Rosa et al.~\cite{de2014online}  &  \surdgreen & \surdgreen  & \crossred & \shortstack{\approxblu \\ \small{(one param.)}} &\surdgreen  & \crossred & \crossred \\
	\hline \rule{0pt}{4ex} Hasan et al.~\cite{hasan2014continuous}  & \shortstack{\approxblu \\ \small{(minibatch)}} & \crossred & \surdgreen & \crossred & \crossred &  \surdgreen  & \crossred \\
	\hline \rule{0pt}{4ex}  Hasan et al.~\cite{hasan2014incremental} &  \shortstack{\approxblu \\ \small{(minibatch)}}  &   \crossred & \crossred & \crossred & \crossred &  \surdgreen & \crossred \\
	\hline \rule{0pt}{4ex}  Hasan et al.~\cite{hasan2015iccv} &  \shortstack{\approxblu \\ \small{(minibatch)}}  &   \crossred & \crossred & \crossred & \crossred &  \surdgreen & \crossred \\ 	
	\hline \rule{0pt}{4ex} FIVER &  \surdgreen   &  \surdgreen &  \surdgreen &  \surdgreen &  \surdgreen &  \surdgreen &  \surdgreen \\
	\hline
\end{tabular}
	\end{center}
	\caption{Identified open challenges in previous work on human action recognition for streaming data. To the best of our knowledge, FIVER is the only algorithm equipped with all these features, crucial for dealing with streaming data. \emph{LEGEND}: \surdgreen: exhibits the feature; \approxblu: partially exhibits the feature; \crossred: does not possess the feature.}
	\vspace{-0.5cm}
	\label{tab:feat_comp}
\end{table*}

A feature tree-based incremental recognition approach was proposed in~\cite{reddy2009incremental}, where the tree is free to grow without bounds as more examples are fed to the learner. As this requires to store all the presented instances, the method is infeasible for continuous recognition from streaming videos where the number of activities can get very large over time. A human tracking-based incremental activity learning framework was brought forward in~\cite{minhas2012incremental}; nevertheless, it requires annotation on the location of the human body in the initial frame, heavily restricting its applicability. For these reasons ~\cite{reddy2009incremental} and~\cite{minhas2012incremental} are not listed in Table~\ref{tab:feat_comp}.

Our work shares similarities with the incremental algorithm proposed in~\cite{de2014online}, on which to some extent we build. Both methods adopt a nonparametric, incremental ball covering of the feature space strategy. FIVER, however, brings to the table crucial new features that makes it uniquely suitable for dealing with streaming data. First, it does not rely on any input parameters,
which are inconvenient to tune in streaming settings.
Second, it limits the model size, thus allowing the tracking of drifting concepts: when the number of allocated balls exceeds a given budget, FIVER discards them with a probability proportional to their error rate. Third, it dynamically adjusts the ball centres, thus yielding very compact models while improving performance. The resulting covering resembles a visual dictionary, learned incrementally and directly usable for predictions (see Fig.~\ref{fig:bow-comp}), where the balls play the role of visual code words. Finally, the active learning module defines the interaction between the learning system and the labeler agent, limiting the number of annotations requested.

The use of incremental active learning for activity recognition tasks has been recently investigated in
\cite{hasan2014continuous,hasan2014incremental}. The approach in~\cite{hasan2014incremental} uses an ensemble of linear SVM classifiers incrementally created in a sequence of mini-batch learning phases. A confidence measure over the SVM outputs is defined, where each individual classifier output is weighted by the training error. Two user-defined thresholds control the query rate of labeled videos.
Non-confident instances, which are close to a class boundary, are forwarded to the annotator, while the others are discarded. Notably, the set of ensemble classifiers can become very large, as an arbitrary number of SVMs  can be added in each batch phase. Furthermore, the method requires model initialisation and has several parameters to be tuned on a validation set; this requirement makes the approach unsuitable for a truly streaming context. The method in~\cite{hasan2014continuous} initially learns features in an unsupervised manner using a deep neural network. Then, a multinomial logistic regression classifier is learned incrementally. The posterior class probability output is used (as in~\cite{hasan2014incremental}) to select what videos need supervised information. This method too presents several parameters, cannot deal with new classes, and requires initialisation. The same authors have recently presented in~\cite{hasan2015iccv} an extension of the methods~\cite{hasan2014incremental,hasan2014continuous} that attempts to mine information from the scene's context. However, the core learning system suffers of the same drawbacks discussed above. Moreover, the active modules used in~\cite{hasan2014incremental,hasan2014continuous,hasan2015iccv} cannot explicitly limit the query rate, a crucial feature for real world applications where the cost of human annotation has to be limited.


\vspace{-0.2cm}
\section{Fast Active Incremental Visual Covering}
\vspace{-0.2cm}
\label{sec:model}
This section describes the FIVER algorithm, which is the heart of our system for activity recognition from streaming videos. Sec.~\ref{sec3.1} describes our incremental visual covering approach based on~\cite{de2014online}, Sec.~\ref{sec:fix_bud} shows how to keep the memory footprint bounded via a technique introduced in~\cite{derosa2015abacoc}, but until now never applied to activity recognition. In Sec.~\ref{sec3.2} we introduce a mechanism that performs active learning on the stream. The FIVER algorithm is finally summarized in Sec.~\ref{sec3.4}.

\vspace{-0.2cm}
\subsection{(Passive) Incremental Learning}
\label{sec3.1}
We assume the learner is trained on a stream of (pre-segmented)\footnote{Although videos are assumed pre-segmented here, it is possible to localise and segment beforehand activity instances, see for example~\cite{hasan2014continuous,hasan2014incremental}.}, labeled videos $(\bV_1,y_1),(\bV_2,y_2),\dots$. Each video $\bV_i$ is associated with a set of $T_i$ local descriptors $\{\bx_t^{(i)}\}_{t=1}^{T_i}$, where each descriptor $\bx_t^{(i)} \in \R^d$ belongs to a $d$-dimensional feature space. Each video label $y_i$ denotes an activity from a set $\sY = \{1,\dots,C\}$ of possible classes, which may change over time. The classifier is trained incrementally, via small adjustments to the current model every time a new labeled video is presented to the learner. In what follows, we drop the superscript $i$ and re-index the local features, thus assuming that the learner is fed a sequence $(\bx_1,y_1),(\bx_2,y_2),\ldots$ of labeled local feature examples, where $(\bx_t,y_t) = \bigl(\bx_t^{(i)},y_i\bigr) \in\R^{d}\times\sY$ for some $t$ and $1 \le t < T_i$. The feature space is adaptively covered by a set $\sS$ of balls depending on the complexity of the classification problem.
Unlike~\cite{de2014online}, where the balls were always centered on input samples, here we adapt the AUTO-ADJ version of ABACOC described in~\cite{derosa2015abacoc}. This uses a $K$-means-like update step in which the centre of each ball is shifted towards the average of the training samples that were correctly predicted by the local classifier; in practise, the balls track the feature clusters, or ``visual codewords''. Following~\cite{derosa2015abacoc} we initialize the radius of any new ball to the distance from the closest ball.
The resulting setting is parameterless\footnote{Unlike~\cite{de2014online}, we set the feature space intrinsic dimension parameter $d$ to 2, as we empirically found that it does not significantly affect performance.}.

\smallskip
\noindent\textbf{Incremental updates.}
The sequence of observed training examples $(\bx_t,y_t)$ is used to build a set $\sS$ of balls that cover the region of the feature space they span. For each ball, an empirical distribution of classes is kept.
%
For each ball centre $\bc_s\in\sS$, we keep updated the number $n_s(y)$ of data points $\bx_t$ of each class $y\in\sY$ that at time $t$ belong to the ball. These counts are used to compute the class probability estimates (activity scores) for each ball centre $\bc_s\in\sS$ as follows:
\vspace{-0.2cm}
\begin{equation}
	\label{eq:prediction}
	p_s(y) = \frac{n_s(y)}{n_s} \qquad y=1,\dots,C
\end{equation}
where $n_s = n_s(1)+\cdots+n_s(C)$.

More specifically, the training algorithm operates as follows ---see Alg.~\ref{alg:train}. Initially, the set of balls $\sS$ is empty. For each training example $\bx_t$, we efficiently\footnote{For example,~\cite{Krauthgamer:2004:NNS:982792.982913} embeds $\sS$ in a tree where nearest neighbour queries and updates can be performed in time $\mathcal{O}(\ln|\sS|)$ ---see also~\cite{conf/nips/KpotufeO13}.} compute its nearest ball centre $\bc_{\pi(t)}\in\sS$. If $\bx_t$ does not belong to its closest ball, i.e., the distance $\rho(\bc_{\pi(t)},\bx_t)$ between $\bx_t$ and $\bc_{\pi(t)}$ is greater than the ball's radius $\ve_{\pi(t)}$, a new ball with centre $\bx_t$ and initial radius $R_t$ equal to $\rho(\bc_{\pi(t)},\bx_t)$ is created and added to $\sS$. The label $y_t$ is used to initialise the empirical class distribution for the new ball via~(\ref{eq:prediction}). If $\bx_t$ does belong to the nearest ball, its label $y_t$ is used to update the error count $m_{\pi(t)}$ for that ball. The local classifier centred at $\bc_{\pi(t)}$ makes a mistake on $(\bx_t,y_t)$ if and only if
$
y_t \neq \argmax_{y\in\sY} p_{\pi(t)}(y)
$.
Whenever this happens, the radius is set to its initial value $R_{\pi(t)}$ scaled by a polynomial function in the current error count: $\ve_{\pi(t)} = R_{\pi(t)}\,m_{\pi(t)}^{-1/4}$.\footnote{This comes from the decay function defined in~\cite{de2014online} with $d=2$.}

If the prediction is correct, the ball centre is set to the average of the correctly classified instances within the ball, allowing it to move towards the majority class centroid. Finally, the class probability estimates $p_{\pi(t)}(y)$  for the local classifier centred in $\bc_{\pi(t)}$ are updated via~(\ref{eq:prediction}). Notably, a-priori knowledge of the full set of classes $\sY$ is not needed, as our incremental learning approach can add new labels to $\sY$ as soon as they first appear in the stream.

\begin{algorithm}                   
	\caption{ABACOC (\cite{derosa2015abacoc}, adapted to feature sets)}      
	\label{alg:train}                   
	\begin{algorithmic}[1]              
		\REQUIRE feature space metric $\rho$ 
		\STATE Ball centres $\sS=\emptyset$ and set of labels $\sY=\emptyset$ initialised
		\FOR{$i=1,2,\dots$}
		\STATE Get labeled video $(\bV_i,y_i)$
		\IF{$y_i \notin \sY$}
		\STATE Set $\sY = \sY \cup \{y_i\}$ // add class on the fly
		\ENDIF
		\FOR{$t=1,\dots,T_i$}
		\IF{$\sS \equiv \emptyset$}
		\IF {$t=1$}
		\STATE save $\bx_1$;
		\ELSE
		\STATE $\sS = \{\bx_2\}$, set  $\ve_2 = R_2 = \rho\bigl(\bx_1,\bx_2)$
		\STATE Use $y_i$ to initialize estimates $p_2$ via (\ref{eq:prediction})
		\ENDIF
		\ELSE	        	
		\STATE Let $\bc_s\in\sS$ be the nearest neighbour of $\bx_t$ in $\sS$
		\IF{$\rho(\bc_s,\bx_t) \le \ve_s$ ($\bx_t$ belongs to ball at $\bc_s$)}
		\IF{${\displaystyle y_i \neq \argmax_{y\in\sY} p_s(y)}$}
		\STATE // shrink radius on errors
		\STATE Set $m_s = m_s + 1$, $\ve_s = R_s\, m_s^{-1/4}$
		\ELSE
		\STATE // update ball centre if correct prediction
		\STATE Set $\Delta=\bx_t-\bc_s$, $n_s=n_s+1$
		\STATE Set $\bc_s=\bc_s+\Delta/n_s$ 					
		\ENDIF
		\STATE Use $y_i$ to update $p_s$ via (\ref{eq:prediction})
		\ELSE
		\STATE $\sS = \sS \cup \{\bx_t\}$, set $\ve_t = R_t = \rho(\bc_s,\bx_t)$, $n_t=1$, $m_t=0$
		\STATE use $y_i$ to initialize estimates $p_t$
		\ENDIF
		\ENDIF	
		\ENDFOR	
		\ENDFOR
	\end{algorithmic}
	
\end{algorithm}

\smallskip
\noindent\textbf{Prediction.}
In the prediction phase, we proceed similarly: for each $\bx_t$ associated with an unlabelled video $\bV_i$, its nearest neighbour $\bc_{\pi(t)}\in\sS$ is efficiently located. Then, assuming that the local features are i.i.d., the label of the test video $\bV_i$ is predicted using the following maximum likelihood estimate:
\begin{equation} \label{eq:pred_vid}
	\yhat_i = \argmax_{y\in\sY} \prod_{t=1}^{T_i} p_{\pi(t)}(y)~.
\end{equation}

\vspace{-0.2cm}
\subsection{Constant Model Size}
\label{sec:fix_bud}
In order to curb the system's memory footprint, we adopt the simple approach proposed in~\cite{derosa2015abacoc} based on deleting existing balls whenever a given budget parameter on the label query rate is attained. This is crucial for real-time applications, as NN search, used in both training and prediction, is logarithmic in the number of balls. The probability of deleting any given ball is proportional to the number of mistakes made so far by the associated classifier.
Namely, if the budget is reached and a new ball has to be added, an existing ball $s$ is deleted with probability
	$\displaystyle \Pr_{\mathrm{disc}}(s) =  \frac{m_s + 1}{\sum_{r \in \sS}{m_r}+|\sS|}$,
where $m_s$ is the number of mistakes made by ball $s \in \sS$. This helps addressing concept drift: ball classifiers that accumulate many mistakes are removed to make room for a more accurate description of the data.

\subsection{Streaming Active Learning} \label{sec:active-learning}
\label{sec3.2}
We now introduce the active learning system for streaming settings, which bounds the rate of queries to human annotators. The technique we propose is inspired from~\cite{zliobaite2014active}.
Whenever a new segmented video is presented to the model, the system makes a prediction and then invokes the active learning module in order to determine whether the label should be requested. In particular, if the confidence of the prediction is below a certain threshold, i.e., the prediction is ambiguous, then a query is issued to the annotator unless the query rate budget is violated. When the label is not requested, the model is not updated. Instead of selecting a fixed confidence threshold on the query instances, we use the so-called Variable Uncertainty Strategy~\cite{zliobaite2014active} (\texttt{VarUnStr}), which queries the least certain instances within a time interval.

\smallskip
\noindent\textbf{Measuring prediction confidence.}
Eq. (\ref{eq:pred_vid}) shows that class estimates $p_{\pi(t)}(y)$ associated with ball centres near the current input instance should be considered more reliable than those associated with faraway centres, as the corresponding region of the feature space has already been explored (see Fig. \ref{fig:spot-gig}-right). We thus adapt the RBF Kernel~\cite{chang2010training} to scale ball estimates based on their distance from the input examples:
\begin{equation}
	\label{eq:pred_conf_weight}
	w_t = \exp\bigg(- \frac{ \rho(\bx_t,\bc_{\pi(t)})^2}{2\epsilon_{\pi(t)}^2}\bigg),
\end{equation}
where the variance is set to the current ball radius $\epsilon_{\pi(t)}$.

Given a test video $\bV_i=\bx_1,\dots,\bx_{T_i}$, we thus define a confidence measure $C_i(y)$ on the estimate of the expected class conditional probability for any given class $y$ as:
\begin{equation} \label{eq:pred_conf}
	C_i(y) = \frac{1}{T_i} \sum_{t=1}^{T_i} w_{t} \, \log p_{\pi(t)}(y) \qquad \forall y \in \mathcal{Y}.
\end{equation}

\noindent\textbf{Updating the confidence threshold}. The \texttt{VarUnStr}~\cite{zliobaite2014active} strategy continuously updates the confidence threshold $\Theta$, which triggers requests for new labels (see Alg.~\ref{alg:vu_strat}). If the prediction confidence is below the current threshold $\Theta$ over the duration of the last observed video, $\Theta$ is decreased by a fraction $\tau$ in order to query the most uncertain instances first. Otherwise, the threshold is increased to avoid interruptions of the learning process when the algorithm is not asking for labels. As explained in~\cite{zliobaite2014active}, the parameter $\tau$ can be set to a default value of $0.01$. In the experimental section we follow this suggestion, thus our algorithm remains parameterless.
\begin{algorithm}[h]                   
	\caption{Variable Uncertainty Strategy}          
	\label{alg:vu_strat}
	\begin{algorithmic}[1]                     
		\REQUIRE incoming video $\bV_i$, classifier \texttt{model}, threshold adjustment step $\tau \in(0,1]$ (default is $0.01$)
		\ENSURE \texttt{labeling} $\in\{true, false\}$
		\STATE \textbf{Initialize:} confidence threshold $\Theta=1$ and store
		the latest value during operation
		\STATE Calculate the confidence associated with the majority class $C_i(\yhat_i)$  via~(\ref{eq:pred_conf})
		\IF {$C_i(\yhat_i)<\Theta$}
		\STATE decrease the confidence threshold $\Theta = (1 - \tau)\Theta$
		\STATE \textbf{return} $true$
		\ELSE
		\STATE increase the confidence threshold $\Theta = (1 + \tau)\Theta$
		\STATE \textbf{return} $false$
		\ENDIF
	\end{algorithmic}
\end{algorithm}


\begin{algorithm}[h]                   
	\caption{FIVER}
	\label{alg:online-budget}
	\begin{algorithmic}[1]                     
		\REQUIRE annotation budget $B$, maximum number of balls $M$, video stream $(\bV_1,y_1),(\bV_2,y_2),\ldots$
		\STATE Initialise online prediction accuracy $A_0 = 0$
		\FOR{$i=1,2,\ldots$}
		\STATE Receive video $\bV_i$
		\STATE Predict $\yhat_i$ (Eq~\ref{eq:pred_vid})
		\STATE Update accuracy $A_i = \bigl(1-\tfrac{1}{i}\bigr)A_{i-1} + \tfrac{1}{i}\mathbb{I}\{\yhat_i = y_i\}$
		\IF {query rate $\leq$ budget $B$}
		\IF {Query Strategy (Alg.~\ref{alg:vu_strat}) returns \emph{true}}
		\STATE Request true label $y_i$ and update query rate
		\STATE Use $(\bV_i,y_i)$ to update model (Alg.~\ref{alg:train})
		\IF {$|S|>M$ (memory exceeded)}
			\STATE Discard one ball (see Sec.~\ref{sec:fix_bud})
	    \ENDIF
		\ENDIF
		\ENDIF
		\ENDFOR
	\end{algorithmic}
\end{algorithm}

\vspace{-0.2cm}
\subsection{FIVER Algorithm}
\label{sec3.4}
The FIVER algorithm (see Alg.~\ref{alg:online-budget}), combines all the elements described above. Namely, FIVER trains the model over the video stream via Alg.~\ref{alg:train}, while controlling the memory footprint as described in Sec.~\ref{sec:fix_bud}. Concurrently, Alg.~\ref{alg:vu_strat} represents the active learning module, which asks only for the most informative instances while not exceeding the budget rate.

\section{Experiments}
\vspace{-0.2cm}
\label{sec:exp}

To emphasize the versatility of our approach, we have tested FIVER
in both the batch and streaming learning settings. In the batch setting,
for each dataset we have followed the standard evaluation protocol
(specific train-test splits or $K$-fold cross-validation with specific values of $K$ ---see below for details) and compared FIVER's results to those of competing incremental and offline methods. In the streaming setting, instead, we assessed different variants of FIVER using the ``online accuracy'' or ``sequential risk''~\cite{gama2013evaluating} as evaluation measure. This measure captures the average error made by the sequence of incrementally learned models in a procedure where we first predict the test item on the current model, and then use the result to adjust the model itself.

Notably, the streaming setting used in our experiments is very strict: we do not use seed training sets, mini-batch training, cross-validation sets, or assume any preliminary knowledge on the number of classes. As the other incremental methods rely on much richer sources of information than those allowed in our streaming setting, we could only evaluate them in the batch setting.



\subsection{Datasets and Feature Extraction}
\label{exp-setup}

We assessed our method on the following datasets:
KTH~\cite{kth} (all scenarios), UCF11~\cite{poppe2010survey} and VIRAT~\cite{oh2011large} for {action recognition}, SKIG~\cite{Liu:2013:LDR:2540128.2540343} and MSRGesture3D~\cite{Wang:2012:RAR:2403006.2403071} for {gesture recognition}, JAPVOW~\cite{japvow} and AUSLAN~\cite{auslan} for {sign language recognition} (UCI Repository~\cite{Asuncion+Newman:2007}).

The first five datasets contain mostly footage material: we decided to extract efficient local features at frame level in order to focus on truly real-time prediction. 
In particular, from KTH, UCF11, and VIRAT sequences we computed improved dense trajectories~\cite{wang2013action}, due to their outstanding performance in action recognition tasks. For each video, three types of features were extracted, namely Histogram of Oriented Gradient (HOG), Histogram of Optical Flow (HOF) and Motion Boundary Histogram (MBH). We ran the code published on the \href{http://lear.inrialpes.fr/people/wang/improved_trajectories}{INRIA Website}, keeping all the default parameters except for the trajectory length, set to 8 frames, and the number of descriptor bins (16 BINs for HOG, MBHx and MBHy, and 18 BINs for HOF). Every 8 frames we obtained a variable number of active trajectories. We then accumulated all the trajectories for each descriptor, and concatenated all the descriptors, obtaining a collection of vectors of 66 dimensions for each video. In this setting, each vector is a summary of the three local descriptors extracted from each video frame. For VIRAT, we initialised the improved trajectory algorithm using the bounding boxes released along with the dataset. For the other datasets, we did not rely on any initialisation. 

From SKIG we extracted the same information as~\cite{Fanello:2013:KSS:2567709.2567745}, which consists of 3DHOF on the RBG frames and GHOG (Global Histogram of Oriented Gradient) on the depth frames. For MSRGesture3D only depth information is available: we extracted two-level pyramidal HOG (PHOG) features using $32$ bins. For all the experiments, the Euclidean distance has been used as metric $\rho$ (see Alg.~\ref{alg:train}), as pilot tests using an $L_{p}$ norm with varying $p$ did not show any improvement.

\begin{table}[h]
	\begin{center}
		\begin{tabular}{|c|c|c|c|c|c|} \hline
			\textsc{dataset} & \textbf{FIVER} & \textbf{Batch} & \textbf{Increm.} \\ \hline\hline
			KTH & \textbf{98.50\%} & $95.00\%$~\cite{wang2011action} & $97.00\%$~\cite{hasan2014incremental} \\ \hline
			UCF11 & \textbf{79.36\%} & $76.10\%$~\cite{liu2009learning} & $66.00\%$~\cite{hasan2014incremental} \\ \hline
			VIRAT &  \textbf{57.20\%} & 55.40\%~\cite{jiang2007towards}  & $54.20\%$~\cite{hasan2014continuous}   \\ \hline
			SKIG & \textbf{98.30\%}  & $88.70\%$~\cite{Liu:2013:LDR:2540128.2540343} &  $97.50\%$~\cite{de2014online}  \\ \hline
			MSRG3D & \textbf{91.25\%} & $86.50\%$~\cite{Wang:2012:RAR:2403006.2403071} &  $90.33\%$~\cite{de2014online}  \\ \hline
			JAPVOW & 96.75\% & $95.67\%$~\cite{antonucci2013b} & \textbf{98.01\%}~\cite{de2014online}  \\ \hline
			AUSLAN & 72.60\% & \textbf{83.81\%}~\cite{antonucci2013b} &  72.32\%~\cite{de2014online}  \\ \hline
		\end{tabular}
	\end{center}
	\caption{Multiclass accuracies of FIVER compared against best batch and incremental methods on seven benchmark datasets.}
	\label{tab:acc_cross}
\end{table}

\subsection{Comparison with competitors in a batch setting}
\label{exp-batch}


We ran a first set of experiments in a batch setting, i.e., running FIVER on a random permutation of the given training set and then applying the resulting classifier on the test set. We compared FIVER against:

\noindent\textbf{Incremental algorithms}~\cite{hasan2014incremental,hasan2014continuous,de2014online}, which follow an incremental learning approach similar to ours. 

\noindent\textbf{Batch algorithms} ~\cite{wang2011action,liu2009learning,jiang2007towards,Liu:2013:LDR:2540128.2540343,Wang:2012:RAR:2403006.2403071,antonucci2013b}, which have unrestricted access to training data for learning, as opposed to incremental methods that can access the data only sequentially. Notably, the performance of incremental algorithms is typically poorer than that obtained using the corresponding batch versions~\cite{dekel2005data}.

We used $5$-fold cross-validation averaged on ten runs for KTH, UCF11 and VIRAT, like the incremental competitors described in Sec.~\ref{sec:rw}. For SKIG and MSRGesture3D we carried out a $3$-fold cross-validation as in~\cite{Wang:2012:RAR:2403006.2403071}. The available training and test sets were used for JAPVOW and AUSLAN. As shown in Table~\ref{tab:acc_cross}, FIVER is among the best methods on all the datasets, no matter whether batch or incremental setting is considered. This demonstrates that our algorithm, combined with state-of-the-art features, provides an accurate and efficient classification system across the board.

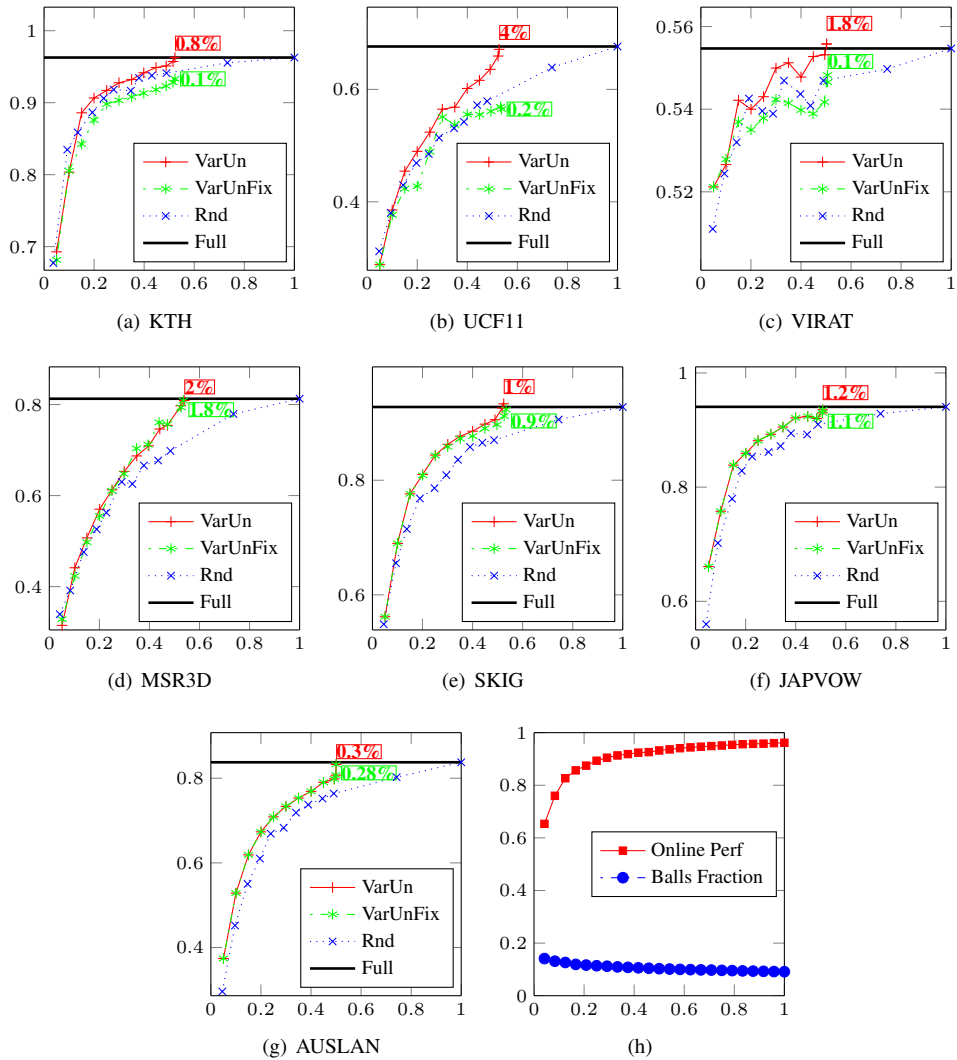
\begin{figure*}[htp!]
	\centering
	\subfigure[KTH]{
		\centering
%
%
\begin{tikzpicture}

\begin{axis}[%
width=0.95092\figurewidthf,
height=\figureheightf,
at={(0\figurewidthf,0\figureheightf)},
scale only axis,
separate axis lines,
every outer x axis line/.append style={black},
every x tick label/.append style={font=\color{black}},
xmin=0,
xmax=1,
every outer y axis line/.append style={black},
every y tick label/.append style={font=\color{black}},
ymin=0.667239844184752,
ymax=1.03271563717307,
legend style={at={(0.97,0.03)},anchor=south east,legend cell align=left,align=left,draw=black}
]
\addplot [color=red,solid,mark size=2.0pt,mark=+,mark options={solid,fill=red}]
  table[row sep=crcr]{%
0.0500834724540902	0.692821368948247\\
0.10016694490818	0.803561491374513\\
0.15025041736227	0.885920979410128\\
0.199777406789093	0.906510851419032\\
0.250417362270451	0.917084028937117\\
0.299387868670006	0.927657206455203\\
0.350027824151363	0.932109070673344\\
0.398441847523651	0.941569282136895\\
0.447412353923205	0.948803561491374\\
0.488035614913745	0.951029493600445\\
0.515303283249861	0.95715080690039\\
0.523094045631608	0.962715637173066\\
};
\addlegendentry{VarUn};

\addplot [color=green,dash pattern=on 1pt off 3pt on 3pt off 3pt,mark size=2.0pt,mark=asterisk,mark options={solid,fill=green}]
  table[row sep=crcr]{%
0.0500834724540902	0.681691708402894\\
0.10016694490818	0.805230940456316\\
0.15025041736227	0.84251530328325\\
0.199777406789093	0.87590428491931\\
0.250417362270451	0.898163606010017\\
0.299387868670006	0.903163481488156\\
0.350027824151363	0.908162437838403\\
0.398441847523651	0.913162705822886\\
0.447412353923205	0.918162537582063\\
0.488035614913745	0.923163581715381\\
0.515303283249861	0.928162978129726\\
0.523094045631608	0.933162216148949\\
};
\addlegendentry{VarUnFix};

\addplot [color=blue,dotted,mark size=2.0pt,mark=x,mark options={solid,fill=blue}]
  table[row sep=crcr]{%
0.0372843628269338	0.677239844184752\\
0.0929326655537006	0.834724540901502\\
0.134668892598776	0.858653311074012\\
0.193099610461881	0.886477462437396\\
0.238174735670562	0.905954368391764\\
0.278241513633834	0.918196994991653\\
0.346688925987757	0.91652754590985\\
0.37952142459655	0.934335002782415\\
0.430717863105175	0.937673900946021\\
0.48914858096828	0.941012799109627\\
0.732888146911519	0.955481357818586\\
1	0.962715637173066\\
};
\addlegendentry{Rnd};

\node[above, right, align=left, inner sep=0mm, font=\bfseries, text=red, draw=red]
at (axis cs:0.523094045631608,0.982715637173066,0) {0.8\%};
\node[below, right, align=left, inner sep=0mm, font=\bfseries, text=green, draw=green]
at (axis cs:0.543094045631608,0.933162216148949,0) {0.1\%};
\addplot [color=black,solid,line width=1.0pt]
  table[row sep=crcr]{%
0	0.962715637173066\\
1	0.962715637173066\\
};
\addlegendentry{Full};

\end{axis}
\end{tikzpicture}%
	}
	\subfigure[UCF11]{
		\centering
%
%
\begin{tikzpicture}

\begin{axis}[%
width=0.95092\figurewidthf,
height=\figureheightf,
at={(0\figurewidthf,0\figureheightf)},
scale only axis,
separate axis lines,
every outer x axis line/.append style={black},
every x tick label/.append style={font=\color{black}},
xmin=0,
xmax=1,
every outer y axis line/.append style={black},
every y tick label/.append style={font=\color{black}},
ymin=0.278259958071279,
ymax=0.746310272536688,
legend style={at={(0.97,0.03)},anchor=south east,legend cell align=left,align=left,draw=black}
]
\addplot [color=red,solid,mark size=2.0pt,mark=+,mark options={solid,fill=red}]
  table[row sep=crcr]{%
0.050314465408805	0.288259958071279\\
0.1	0.385324947589099\\
0.150314465408805	0.454507337526205\\
0.2	0.489727463312369\\
0.250314465408805	0.523899371069183\\
0.30020964360587	0.564779874213836\\
0.349895178197065	0.568343815513627\\
0.39979035639413	0.60167714884696\\
0.448427672955975	0.616142557651992\\
0.491194968553459	0.635639412997904\\
0.522851153039832	0.659329140461216\\
0.527253668763103	0.671310272536688\\
};
\addlegendentry{VarUn};

\addplot [color=green,dash pattern=on 1pt off 3pt on 3pt off 3pt,mark size=2.0pt,mark=asterisk,mark options={solid,fill=green}]
  table[row sep=crcr]{%
0.050314465408805	0.288259958071279\\
0.10020964360587	0.376519916142558\\
0.150314465408805	0.423270440251572\\
0.20020964360587	0.427672955974843\\
0.250314465408805	0.488993710691824\\
0.30041928721174	0.550943396226415\\
0.349685534591195	0.536897274633124\\
0.4	0.555555555555556\\
0.449056603773585	0.555136268343816\\
0.494339622641509	0.561006289308176\\
0.532704402515723	0.568972746331237\\
0.536897274633124	0.564989517819707\\
};
\addlegendentry{VarUnFix};

\addplot [color=blue,dotted,mark size=2.0pt,mark=x,mark options={solid,fill=blue}]
  table[row sep=crcr]{%
0.0480083857442348	0.311949685534591\\
0.0943396226415095	0.380293501048218\\
0.143605870020964	0.429769392033543\\
0.19685534591195	0.468972746331237\\
0.245911949685535	0.485115303983229\\
0.287211740041929	0.514046121593291\\
0.347169811320755	0.531027253668763\\
0.386582809224319	0.542348008385744\\
0.439412997903564	0.572117400419287\\
0.479454926624738	0.579035639412998\\
0.737945492662474	0.638993710691824\\
1	0.676310272536688\\
};
\addlegendentry{Rnd};

\node[above, right, align=left, inner sep=0mm, font=\bfseries, text=red, draw=red]
at (axis cs:0.527253668763103,0.701310272536688,0) {4\%};
\node[below, right, align=left, inner sep=0mm, font=\bfseries, text=green, draw=green]
at (axis cs:0.556897274633124,0.564989517819707,0) {0.2\%};
\addplot [color=black,solid,line width=1.0pt]
  table[row sep=crcr]{%
0	0.676310272536688\\
1	0.676310272536688\\
};
\addlegendentry{Full};

\end{axis}
\end{tikzpicture}%
	}
	\subfigure[VIRAT]{
		\centering
%
%
\begin{tikzpicture}

\begin{axis}[%
width=0.95092\figurewidthf,
height=\figureheightf,
at={(0\figurewidthf,0\figureheightf)},
scale only axis,
separate axis lines,
every outer x axis line/.append style={black},
every x tick label/.append style={font=\color{black}},
xmin=0,
xmax=1,
every outer y axis line/.append style={black},
every y tick label/.append style={font=\color{black}},
ymin=0.501024643320363,
ymax=0.564690877648076,
legend style={at={(0.97,0.03)},anchor=south east,legend cell align=left,align=left,draw=black}
]
\addplot [color=red,solid,mark size=2.0pt,mark=+,mark options={solid,fill=red}]
  table[row sep=crcr]{%
0.0505836575875486	0.521184608733247\\
0.100518806744488	0.526588845654993\\
0.150453955901427	0.542153047989624\\
0.200389105058366	0.539991353220925\\
0.250324254215305	0.543017725897103\\
0.300043233895374	0.549935149156939\\
0.349978383052313	0.551232166018158\\
0.399913532209252	0.54777345438824\\
0.449200172935582	0.552745352356247\\
0.495028102031993	0.553177691309987\\
0.502377864245569	0.555771725032425\\
0.502161694768699	0.555771725032425\\
};
\addlegendentry{VarUn};

\addplot [color=green,dash pattern=on 1pt off 3pt on 3pt off 3pt,mark size=2.0pt,mark=asterisk,mark options={solid,fill=green}]
  table[row sep=crcr]{%
0.0505836575875486	0.521184608733247\\
0.100518806744488	0.527885862516213\\
0.150453955901427	0.536964980544747\\
0.200172935581496	0.535019455252918\\
0.250324254215305	0.537829658452227\\
0.300043233895374	0.542369217466494\\
0.349978383052313	0.541504539559014\\
0.399913532209252	0.539775183744055\\
0.448335495028102	0.538910505836576\\
0.494163424124514	0.541720709035884\\
0.505620406398617	0.548261997405966\\
0.506052745352356	0.546532641591007\\
};
\addlegendentry{VarUnFix};

\addplot [color=blue,dotted,mark size=2.0pt,mark=x,mark options={solid,fill=blue}]
  table[row sep=crcr]{%
0.0473411154345006	0.511024643320363\\
0.0940337224383917	0.524427150886295\\
0.142888024210981	0.53199308257674\\
0.191958495460441	0.542585386943364\\
0.245352356247298	0.539559014267185\\
0.288370082144401	0.538910505836576\\
0.332036316472114	0.546908776480761\\
0.397751837440553	0.543666234327713\\
0.437959360138348	0.540856031128405\\
0.489407695633377	0.546908776480761\\
0.744271508862949	0.549718979680069\\
1	0.554690877648076\\
};
\addlegendentry{Rnd};

\node[above, right, align=left, inner sep=0mm, font=\bfseries, text=red, draw=red]
at (axis cs:0.502161694768699,0.560771725032425,0) {1.8\%};
\node[below, right, align=left, inner sep=0mm, font=\bfseries, text=green, draw=green]
at (axis cs:0.506052745352356,0.551532641591007,0) {0.1\%};
\addplot [color=black,solid,line width=1.0pt]
  table[row sep=crcr]{%
0	0.554690877648076\\
1	0.554690877648076\\
};
\addlegendentry{Full};

\end{axis}
\end{tikzpicture}%
	}
	\subfigure[MSR3D]{
		\centering
%
%
\begin{tikzpicture}

\begin{axis}[%
width=0.95092\figurewidthf,
height=\figureheightf,
at={(0\figurewidthf,0\figureheightf)},
scale only axis,
separate axis lines,
every outer x axis line/.append style={black},
every x tick label/.append style={font=\color{black}},
xmin=0,
xmax=1,
every outer y axis line/.append style={black},
every y tick label/.append style={font=\color{black}},
ymin=0.304814814814815,
ymax=0.882812812812813,
legend style={at={(0.97,0.03)},anchor=south east,legend cell align=left,align=left,draw=black}
]
\addplot [color=red,solid,mark size=2.0pt,mark=+,mark options={solid,fill=red}]
  table[row sep=crcr]{%
0.0510510510510511	0.314814814814815\\
0.102102102102102	0.441441441441441\\
0.15015015015015	0.507007007007007\\
0.201201201201201	0.57007007007007\\
0.250750750750751	0.612612612612613\\
0.299299299299299	0.653153153153153\\
0.349349349349349	0.687187187187187\\
0.397397397397397	0.709209209209209\\
0.441441441441441	0.746246246246246\\
0.471971971971972	0.754254254254254\\
0.525025025025025	0.797297297297297\\
0.539039039039039	0.808808808808809\\
};
\addlegendentry{VarUn};

\addplot [color=green,dash pattern=on 1pt off 3pt on 3pt off 3pt,mark size=2.0pt,mark=asterisk,mark options={solid,fill=green}]
  table[row sep=crcr]{%
0.051051051051051	0.328328328328328\\
0.102102102102102	0.423423423423423\\
0.15015015015015	0.497497497497498\\
0.201201201201201	0.555555555555556\\
0.25025025025025	0.610610610610611\\
0.298298298298298	0.647647647647648\\
0.348348348348348	0.703703703703704\\
0.394394394394394	0.711711711711712\\
0.438438438438438	0.760760760760761\\
0.472472472472472	0.756756756756757\\
0.526526526526527	0.793793793793794\\
0.535535535535536	0.808808808808809\\
};
\addlegendentry{VarUnFix};

\addplot [color=blue,dotted,mark size=2.0pt,mark=x,mark options={solid,fill=blue}]
  table[row sep=crcr]{%
0.042042042042042	0.339339339339339\\
0.0835835835835836	0.391391391391391\\
0.137637637637638	0.475975975975976\\
0.189189189189189	0.526026026026026\\
0.228228228228228	0.562562562562563\\
0.289289289289289	0.63013013013013\\
0.331831831831832	0.625625625625626\\
0.377877877877878	0.666166166166166\\
0.434934934934935	0.677177177177177\\
0.483983983983984	0.698198198198198\\
0.735735735735736	0.779279279279279\\
1	0.812812812812813\\
};
\addlegendentry{Rnd};

\node[above, right, align=left, inner sep=0mm, font=\bfseries, text=red, draw=red]
at (axis cs:0.539039039039039,0.838808808808809,0) {2\%};
\node[below, right, align=left, inner sep=0mm, font=\bfseries, text=green, draw=green]
at (axis cs:0.555535535535536,0.788808808808809,0) {1.8\%};
\addplot [color=black,solid,line width=1.0pt]
  table[row sep=crcr]{%
0	0.812812812812813\\
1	0.812812812812813\\
};
\addlegendentry{Full};

\end{axis}
\end{tikzpicture}%
	}
	\subfigure[SKIG]{
		\centering
%
%
\begin{tikzpicture}

\begin{axis}[%
width=0.95092\figurewidthf,
height=\figureheightf,
at={(0\figurewidthf,0\figureheightf)},
scale only axis,
separate axis lines,
every outer x axis line/.append style={black},
every x tick label/.append style={font=\color{black}},
xmin=0,
xmax=1,
every outer y axis line/.append style={black},
every y tick label/.append style={font=\color{black}},
ymin=0.538765432098765,
ymax=0.997777777777778,
legend style={at={(0.97,0.03)},anchor=south east,legend cell align=left,align=left,draw=black}
]
\addplot [color=red,solid,mark size=2.0pt,mark=+,mark options={solid,fill=red}]
  table[row sep=crcr]{%
0.0503086419753086	0.561728395061728\\
0.100308641975309	0.689814814814815\\
0.150308641975309	0.776234567901235\\
0.200308641975309	0.809876543209877\\
0.250925925925926	0.843827160493827\\
0.300925925925926	0.862654320987654\\
0.350617283950617	0.876851851851852\\
0.400308641975309	0.885493827160494\\
0.449382716049383	0.89783950617284\\
0.489506172839506	0.905555555555556\\
0.51820987654321	0.925925925925926\\
0.524382716049383	0.933333333333333\\
};
\addlegendentry{VarUn};

\addplot [color=green,dash pattern=on 1pt off 3pt on 3pt off 3pt,mark size=2.0pt,mark=asterisk,mark options={solid,fill=green}]
  table[row sep=crcr]{%
0.0503086419753086	0.561728395061728\\
0.100308641975309	0.689814814814815\\
0.150308641975309	0.776234567901235\\
0.200617283950617	0.808641975308642\\
0.250925925925926	0.84320987654321\\
0.300925925925926	0.858333333333333\\
0.350925925925926	0.872530864197531\\
0.400308641975309	0.876851851851852\\
0.449074074074074	0.890432098765432\\
0.497530864197531	0.896913580246914\\
0.526234567901235	0.912037037037037\\
0.533950617283951	0.923456790123457\\
};
\addlegendentry{VarUnFix};

\addplot [color=blue,dotted,mark size=2.0pt,mark=x,mark options={solid,fill=blue}]
  table[row sep=crcr]{%
0.0450617283950617	0.548765432098765\\
0.0941358024691358	0.655246913580247\\
0.137345679012346	0.714814814814815\\
0.189506172839506	0.76820987654321\\
0.247839506172839	0.786111111111111\\
0.295061728395062	0.808950617283951\\
0.341666666666667	0.835802469135802\\
0.389814814814815	0.858024691358025\\
0.438888888888889	0.865432098765432\\
0.485802469135802	0.870061728395062\\
0.743518518518519	0.905864197530864\\
1	0.927777777777778\\
};
\addlegendentry{Rnd};

\node[above, right, align=left, inner sep=0mm, font=\bfseries, text=red, draw=red]
at (axis cs:0.524382716049383,0.963333333333333,0) {1\%};
\node[below, right, align=left, inner sep=0mm, font=\bfseries, text=green, draw=green]
at (axis cs:0.553950617283951,0.903456790123457,0) {0.9\%};
\addplot [color=black,solid,line width=1.0pt]
  table[row sep=crcr]{%
0	0.927777777777778\\
1	0.927777777777778\\
};
\addlegendentry{Full};

\end{axis}
\end{tikzpicture}%
	}
	\subfigure[JAPVOW]{
		\centering
%
%
\begin{tikzpicture}

\begin{axis}[%
width=0.95092\figurewidthf,
height=\figureheightf,
at={(0\figurewidthf,0\figureheightf)},
scale only axis,
separate axis lines,
every outer x axis line/.append style={black},
every x tick label/.append style={font=\color{black}},
xmin=0,
xmax=1,
every outer y axis line/.append style={black},
every y tick label/.append style={font=\color{black}},
ymin=0.549895833333333,
ymax=1.010625,
legend style={at={(0.97,0.03)},anchor=south east,legend cell align=left,align=left,draw=black}
]
\addplot [color=red,solid,mark size=2.0pt,mark=+,mark options={solid,fill=red}]
  table[row sep=crcr]{%
0.0515625	0.6609375\\
0.100520833333333	0.7578125\\
0.151041666666667	0.838541666666667\\
0.200520833333333	0.859375\\
0.249479166666667	0.88125\\
0.300520833333333	0.892708333333333\\
0.348958333333333	0.905208333333333\\
0.399479166666667	0.921354166666667\\
0.447916666666667	0.923958333333333\\
0.483854166666667	0.9203125\\
0.506770833333333	0.930208333333333\\
0.5078125	0.935416666666667\\
};
\addlegendentry{VarUn};

\addplot [color=green,dash pattern=on 1pt off 3pt on 3pt off 3pt,mark size=2.0pt,mark=asterisk,mark options={solid,fill=green}]
  table[row sep=crcr]{%
0.0515625	0.6609375\\
0.100520833333333	0.7578125\\
0.151041666666667	0.838541666666667\\
0.200520833333333	0.859375\\
0.249479166666667	0.88125\\
0.300520833333333	0.892708333333333\\
0.348958333333333	0.905208333333333\\
0.399479166666667	0.921354166666667\\
0.447916666666667	0.923958333333333\\
0.483854166666667	0.9203125\\
0.506770833333333	0.930208333333333\\
0.5078125	0.935416666666667\\
};
\addlegendentry{VarUnFix};

\addplot [color=blue,dotted,mark size=2.0pt,mark=x,mark options={solid,fill=blue}]
  table[row sep=crcr]{%
0.0421875	0.559895833333333\\
0.0890625	0.702083333333333\\
0.145833333333333	0.7796875\\
0.184375	0.828645833333333\\
0.225520833333333	0.853645833333333\\
0.291145833333333	0.861458333333333\\
0.339583333333333	0.871875\\
0.38125	0.894270833333333\\
0.445833333333333	0.8921875\\
0.4890625	0.909895833333333\\
0.738541666666667	0.928645833333333\\
1	0.940625\\
};
\addlegendentry{Rnd};

\node[above, right, align=left, inner sep=0mm, font=\bfseries, text=red, draw=red]
at (axis cs:0.5078125,0.965416666666667,0) {1.2\%};
\node[below, right, align=left, inner sep=0mm, font=\bfseries, text=green, draw=green]
at (axis cs:0.5278125,0.915416666666667,0) {1.1\%};
\addplot [color=black,solid,line width=1.0pt]
  table[row sep=crcr]{%
0	0.940625\\
1	0.940625\\
};
\addlegendentry{Full};

\end{axis}
\end{tikzpicture}%
	}
	\subfigure[AUSLAN]{
		\centering
%
%
\begin{tikzpicture}

\begin{axis}[%
width=0.95092\figurewidthf,
height=\figureheightf,
at={(0\figurewidthf,0\figureheightf)},
scale only axis,
separate axis lines,
every outer x axis line/.append style={black},
every x tick label/.append style={font=\color{black}},
xmin=0,
xmax=1,
every outer y axis line/.append style={black},
every y tick label/.append style={font=\color{black}},
ymin=0.286166341780377,
ymax=0.907686809616634,
legend style={at={(0.97,0.03)},anchor=south east,legend cell align=left,align=left,draw=black}
]
\addplot [color=red,solid,mark size=2.0pt,mark=+,mark options={solid,fill=red}]
  table[row sep=crcr]{%
0.0502923976608187	0.374009096816114\\
0.100194931773879	0.528525016244315\\
0.15009746588694	0.618843404808317\\
0.200129954515919	0.673294346978557\\
0.250162443144899	0.708771929824562\\
0.29993502274204	0.733593242365172\\
0.35009746588694	0.753216374269006\\
0.400259909031839	0.768940870695257\\
0.449772579597141	0.790383365821962\\
0.492917478882391	0.798310591293047\\
0.499675113710202	0.807277452891488\\
0.500194931773879	0.832686809616634\\
};
\addlegendentry{VarUn};

\addplot [color=green,dash pattern=on 1pt off 3pt on 3pt off 3pt,mark size=2.0pt,mark=asterisk,mark options={solid,fill=green}]
  table[row sep=crcr]{%
0.0502923976608187	0.374009096816114\\
0.100194931773879	0.528525016244315\\
0.15009746588694	0.618843404808317\\
0.200129954515919	0.673294346978557\\
0.250162443144899	0.708771929824562\\
0.29993502274204	0.733593242365172\\
0.35009746588694	0.753216374269006\\
0.400259909031839	0.768940870695257\\
0.449772579597141	0.790383365821962\\
0.492917478882391	0.798310591293047\\
0.499675113710202	0.807277452891488\\
0.500194931773879	0.832686809616634\\
};
\addlegendentry{VarUnFix};

\addplot [color=blue,dotted,mark size=2.0pt,mark=x,mark options={solid,fill=blue}]
  table[row sep=crcr]{%
0.0457439896036387	0.296166341780377\\
0.0965562053281351	0.452111760883691\\
0.146718648473034	0.550357374918778\\
0.195971410006498	0.610136452241715\\
0.239246263807667	0.669005847953216\\
0.29083820662768	0.682910981156595\\
0.339961013645224	0.718778427550357\\
0.389083820662768	0.737491877842755\\
0.445873944119558	0.751916829109812\\
0.4909681611436	0.763872644574399\\
0.741910331384016	0.80233918128655\\
1	0.837686809616634\\
};
\addlegendentry{Rnd};

\node[above, right, align=left, inner sep=0mm, font=\bfseries, text=red, draw=red]
at (axis cs:0.500194931773879,0.862686809616634,0) {0.3\%};
\node[below, right, align=left, inner sep=0mm, font=\bfseries, text=green, draw=green]
at (axis cs:0.520194931773879,0.812686809616634,0) {0.28\%};
\addplot [color=black,solid,line width=1.0pt]
  table[row sep=crcr]{%
0	0.837686809616634\\
1	0.837686809616634\\
};
\addlegendentry{Full};

\end{axis}
\end{tikzpicture}%
	}
	\subfigure[]{
		\centering
%
%
\begin{tikzpicture}

\begin{axis}[%
width=0.95092\figurewidthf,
height=\figureheightf,
at={(0\figurewidthf,0\figureheightf)},
scale only axis,
separate axis lines,
every outer x axis line/.append style={black},
every x tick label/.append style={font=\color{black}},
xmin=0,
xmax=1,
every outer y axis line/.append style={black},
every y tick label/.append style={font=\color{black}},
ymin=0,
ymax=1,
legend style={at={(0.97,0.5)},anchor=east,legend cell align=left,align=left,draw=black}
]
\addplot [color=red,solid,mark size=1.4pt,mark=square*,mark options={solid,fill=red}]
  table[row sep=crcr]{%
0.0416666666666667	0.653333333333333\\
0.0833333333333333	0.76\\
0.125	0.826666666666667\\
0.166666666666667	0.856666666666667\\
0.208333333333333	0.874666666666667\\
0.25	0.893333333333333\\
0.291666666666667	0.90476\\
0.333333333333333	0.913333333333333\\
0.375	0.91852\\
0.416666666666667	0.924\\
0.458333333333333	0.92606\\
0.5	0.93222\\
0.541666666666667	0.93641\\
0.583333333333333	0.940953333333333\\
0.625	0.944\\
0.666666666666667	0.946666666666667\\
0.708333333333333	0.949016666666667\\
0.75	0.951113333333333\\
0.791666666666667	0.953683333333333\\
0.833333333333333	0.956\\
0.875	0.95746\\
0.916666666666667	0.958183333333333\\
0.958333333333333	0.96\\
1	0.9616\\
};
\addlegendentry{Online Perf};

\addplot [color=blue,dash pattern=on 1pt off 3pt on 3pt off 3pt,mark size=2.0pt,mark=*,mark options={solid,fill=blue}]
  table[row sep=crcr]{%
0.0416666666666667	0.14116368860912\\
0.0833333333333333	0.131259694364501\\
0.125	0.126298005587038\\
0.166666666666667	0.119049751093259\\
0.208333333333333	0.116416201166324\\
0.25	0.114002029582308\\
0.291666666666667	0.112124194312434\\
0.333333333333333	0.109559130689236\\
0.375	0.107767520445597\\
0.416666666666667	0.106005182753398\\
0.458333333333333	0.104362571449878\\
0.5	0.102742573239648\\
0.541666666666667	0.101368952392592\\
0.583333333333333	0.100158132627162\\
0.625	0.0991112457302168\\
0.666666666666667	0.0981852191326383\\
0.708333333333333	0.0974325989276772\\
0.75	0.0964357214626443\\
0.791666666666667	0.0954045845603636\\
0.833333333333333	0.0943501910817316\\
0.875	0.0935190937571415\\
0.916666666666667	0.0926402322423536\\
0.958333333333333	0.0918041684678613\\
1	0.0912485323504941\\
};
\addlegendentry{Balls Fraction};

\end{axis}
\end{tikzpicture}%
	}
	
	\caption{
		Plots ({a}) to ({g}) show the active online performance of the Full, Rnd, VarUn and VarUnFix variants of FIVER on different benchmarks. The x-axis is the percentage of label requested by the active learning module, while the y-axis plots the average online accuracy over ten random permutations of the videos. In the coloured boxes, the percentage of the input data selected as centres by \textit{VarUn}(red) and \textit{VarUnFix}(green) with budget $B=1$ are shown. Plot \textit{h} represents the evolution of accuracy and model size over the sequentially fed videos on the KTH dataset. The blu (circle) curve shows the fraction of input data selected as centres, and the red (square) curve the online accuracy. Notably, the fraction of centers added diminish over time as the accuracy improves.
	}
	\label{fig:perf-seq}\vspace{-3mm}
\end{figure*}
\begin{figure*}[ht!]
	\centering
	\includegraphics[width=1\textwidth]{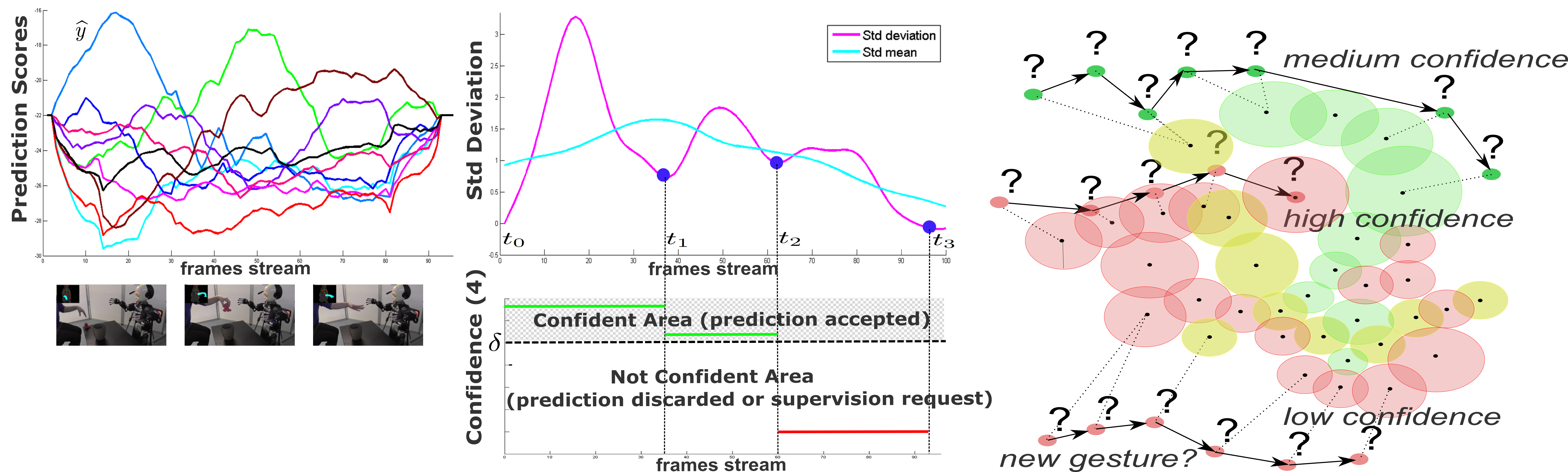}
	\caption{%
		Left: evolution of 10 action class probabilities over time for a test sequence containing three actions. Middle top: the pink line plots the standard deviation of class probabilities for each frame of the same sequence ---the cyan curve is their average standard deviation computed over a short interval of frames. For each segmented activity, a confidence measure is computed (\ref{eq:pred_conf}) ---the predicted activity label is discarded when confidence is below an adaptive threshold $\Theta$. Right: examples of three time series associated with low, medium and high confidence, respectively.
	}%
	\label{fig:spot-gig}
	\vspace{-3mm}
\end{figure*}
\subsection{FIVER performance in a streaming setting}
\label{exp-online}

In a pure streaming setting the data arrive sequentially, and the number of activities depicted in each video is not known a priori. We conducted extensive experiments exploring the following scenarios, which correspond to different variants of FIVER:

\noindent\textbf{\textit{Full.}} This is the least realistic case, where we assume that all the incoming instances are manually annotated, we have no memory requirements, and we use each incoming training sample to incrementally update the model.

\noindent\textbf{\textit{VarUn.}} In this case, the active learning component described in Alg.~\ref{alg:online-budget} is used, including the \texttt{VarUnStr} strategy described in Alg.~\ref{alg:vu_strat}, to decide what instances require manual annotation. The query rate, calculated as the fraction of videos for which a label was requested among those observed so far \cite{zliobaite2014active}, is upper bounded by an input budget parameter $B\in(0,1]$.

\noindent\textbf{\textit{Rnd.}} The Random strategy queries the labels of incoming instances with probability equal to the query rate budget $B$.

\noindent\textbf{\textit{VarUnFix.}} In this very realistic scenario we also assume that we have a limited memory space to store labeled training instances. We apply the method of Sec.~\ref{sec:fix_bud} to limit the number of balls stored in the model. In all our tests we set the model size to 5000 instances. 
Notably, in BoW methods a large amount of codewords are generally necessary to successfully predict video labels ---see for instance~\cite{wang2013action}, where the authors use four different visual vocabularies of 100,000 words (one for each local descriptor).


We built ten random permutations of the videos in each dataset. The algorithm had to predict the label of each new incoming video. After each prediction, if the active learning system requested the true label, the video along with its label were fed to the model as a new training example. We ran all the competing algorithms with the same range $B\in\{.05,.1,.15,.2,.25,.3,.35,.4,.45,.5,.75,1\}$ of budget values, and plotted the resulting online accuracy, averaged over ten different streams, against the average query rate. Importantly, the budget is only an upper limit to the actual query rate ---algorithms generally ask for a smaller number of annotations.

Note that FIVER does not need any validation set as it has no parameters to tune. This is very important in the streaming context, where non-adaptive methods which tune their parameters in an initial validation stage may perform suboptimally on future data. Plots \textit{a} to \textit{g} in Fig.~\ref{fig:perf-seq} represent the recorded performance on the various benchmarks for all the presented scenarios. The figure shows that \textit{VarUn} performs as well as \textit{Full} on most datasets, even though it queries only around 50\% of all the labels. On KTH, for example, \textit{VarUn} achieves 90\% online accuracy while accessing only less than 20\% of the labels. The \textit{Rnd} method performs typically worse and needs all the labels to reach the performance of \textit{Full}. \textit{VarUnFix} works almost as well as \textit{VarUn} on the simplest datasets and slightly worse on the complex ones; this is due to the fixed budget control that has to discard information in order to keep the model size fixed. For example, both \textit{VarUn} and \textit{Rnd} use $4\%$ of the input data around the $50\%$ query rate for UCF11, whereas \textit{VarUnFix} use only $0.2\%$ of the data --this is shown in the red and green boxes in Fig.~\ref{fig:perf-seq} respectively, as final percentage of input examples used as model centres. Therefore, \textit{VarUnFix} is extremely good at compressing the data, and allows for efficient computation at the cost of a little performance degradation.

\subsection{Active Continuous Activity Recognition}
\label{exp-stream}

Although in Sec.~\ref{sec:model} we assumed that the incoming videos $V_i$ are pre-segmented, whenever feature vectors $\{\bx_t^{(i)}\}_{t=1}^{T_i}$ are extracted on a frame-by-frame basis we can exploit the activity scores~(\ref{eq:pred_vid}) computed over a short temporal window to perform automated temporal segmentation. This segmentation procedure is based on the evolution of class probabilities over time ~\cite{de2014online} (Fig.~\ref{fig:spot-gig}.Left), where transitions between action instances can be associated with local minima of the standard deviation of class scores (pink curve) over the temporal window (Fig. ~\ref{fig:spot-gig}.Middle-top).

In addition, unlike what has been done in~\cite{de2014online}, we use the confidence measure~(\ref{eq:pred_conf}) to discard or send to supervision any detected activity with confidence below a certain threshold (Fig.~\ref{fig:spot-gig},Middle-bottom), as discussed in Sec.~\ref{sec:active-learning}. This is crucial in applications such as human-robot interaction, where it is preferable for the robot not to perform any action when prediction confidence is low, as this may lead to safety issues or communication errors.

We tested this active approach to temporal segmentation on the same dataset of ten manipulative actions used in~\cite{de2014online}. Each action was recorded $60$ times in two different illumination settings and backgrounds, and 3DHOF and HOG descriptors were extracted for each frame.
We excluded four out of ten gestures from the learning phase, and evaluated our algorithm on sequences representing pick and place activities formed by grasping, moving and releasing actions. The system was evaluated on its ability to predict the correct class when a known gesture was performed, and to request supervision when an unknown gesture was observed.

To compare the estimated class sequence with the ground truth we employed the Levenshtein distance~\cite{levi66}, originally used in~\cite{de2014online}:
$
\tfrac{S+D+I}{N}
$.
In this case, each action is treated as a symbol in a sequence -- S represents the number of substitutions (misclassifications), D the number of deletions (false negatives) and I the number of insertions (false positives). Over $20$ test sequences, we achieved a Levenshtein distance error of $0.14$, compared to the $0.36$ reported in~\cite{de2014online}.

\vspace{-0.2cm}
\section{Conclusion and future work}
\label{sec:conc}
We presented an incremental active human activity recognition framework, well suited for streaming recognition problems, especially when the amount of data to process is large. Our approach exhibits a number of desirable features: it deals with sets of local descriptors extracted from videos, it learns in an incremental fashion, it embeds an active learning module, it is capable of learning new classes on the fly, it limits memory usage, and it predicts new data in real-time. In addition, the method is nonparametric and does not require expensive validation sessions for training, as it has no parameters to be tuned. Results demonstrate its competitiveness in terms of accuracy with respect to traditional batch approaches, as well as promising performance in a truly streaming scenario. Future research will explore the use of confidence measures to automatically discover new activity classes by associating them with low confidence trajectories (see Fig.~\ref{fig:spot-gig}, right).




\end{document}